\documentclass[11pt]{article}

\usepackage[utf8]{inputenc}
\usepackage[T1]{fontenc}
\usepackage{times}
\usepackage{geometry}
\usepackage{amsmath, amssymb}
\usepackage{graphicx}
\usepackage{enumitem}
\usepackage{url}
\usepackage{natbib}
\usepackage{titlesec}
\usepackage{hyperref}
\usepackage{xcolor}
\usepackage{booktabs}
\hypersetup{
    colorlinks=true,
    linkcolor=blue,
    citecolor=blue,
    urlcolor=blue
}

\titleformat{\section}[block]{\normalfont\Large\bfseries}{\thesection.}{1em}{}
\titleformat{\subsection}[block]{\normalfont\large\bfseries}{\thesubsection.}{1em}{}

\title{Majority Rules: LLM Ensemble is a Winning Approach for Content Categorization}
\author{
  Ariel Kamen \\ RingCentral Inc. \\ \texttt{ariel.kamen@ringcentral.com}
  \and
  Yakov Kamen \\ Relevad Corporation \\ \texttt{yakov@relevad.com}
}
\date{November 1, 2025}

\begin{document}

\maketitle

\begin{abstract}
This study introduces an ensemble framework for unstructured text categorization using large language models (LLMs). By integrating multiple models, the ensemble large language model (eLLM) framework addresses common weaknesses of individual systems, including inconsistency, hallucination, category inflation, and misclassification. The eLLM approach yields a substantial performance improvement of up to 65\% in F1-score over the strongest single model. We formalize the ensemble process through a mathematical model of collective decision-making and establish principled aggregation criteria. Using the Interactive Advertising Bureau (IAB) hierarchical taxonomy, we evaluate ten state-of-the-art LLMs under identical zero-shot conditions on a human-annotated corpus of 8{,}660 samples. Results show that individual models plateau in performance due to the compression of semantically rich text into sparse categorical representations, while eLLM improves both robustness and accuracy. With a diverse consortium of models, eLLM achieves near human-expert-level performance, offering a scalable and reliable solution for taxonomy-based classification that may significantly reduce dependence on human expert labeling.
\end{abstract}


\section{Introduction}\label{sec:introduction}

Text categorization (also termed text classification) is a foundational problem in natural language processing. Early systems relied on human experts to assign categories using controlled vocabularies and taxonomies. This labor-intensive workflow gradually gave way to rule-based heuristics and, later, to classical machine-learning pipelines with domain-specific training and hand-engineered features. While effective within constrained settings, these approaches required substantial manual effort and did not scale easily across heterogeneous domains.

Large language models (LLMs)—including GPT, Gemini, Claude, Grok, LLaMA, Mistral, xAI, and DeepSeek—have introduced strong zero-shot capabilities, motivating efforts to replace task-specific classifiers with general-purpose LLMs. Despite notable successes of LLMs in reasoning, scientific discovery, and content generation, their performance in taxonomy-based categorization has lagged. Empirically, single models exhibit instability, category inflation, and hallucination, often producing incoherent or non-existent labels.

This performance gap reflects the unique nature of categorization as an information-compression task: semantically rich, unstructured inputs must be mapped into a small set of category names defined by a fixed taxonomy. Such taxonomies typically represent the real world sparsely, and individual texts may map to multiple categories with varying levels of relevance. Accurate mapping frequently requires background knowledge external to the text—such as domain conventions or factual context—that distinguishes professional experts from general readers. In our experiments with ten widely used zero-shot LLMs spanning multiple families and release generations, we observe a consistent performance plateau: increasing model size or algorithmic sophistication does not yield commensurate gains once a threshold is reached \citep{kamen2025order}.

To address these limitations, we developed an ensemble of LLMs (eLLM), treating each model as an independent expert and combining their predictions through a collective decision-making rule. This design is motivated by the well-established benefits of aggregation in both human and algorithmic committees, where diversity and redundancy suppress individual errors and reduce variance. By leveraging models built on diverse architectures, training paradigms, and knowledge bases, the ensemble extends the effective knowledge universe of any single LLM, enabling more comprehensive coverage of sparse taxonomies. In taxonomy classification, we find that ensembling mitigates common single-model failure modes—such as hallucinations and inconsistent label usage—while substantially improving both accuracy and stability. The results are striking: ensembles with sufficient diversity not only surpass the strongest individual models but, in some conditions, approach or exceed the consistency of human expert annotation. The principal trade-off lies in computational cost, since multiple model evaluations are required at inference time. This direction aligns with growing evidence that collaborative or multi-agent LLM strategies can overcome inherent limitations of single-model systems \citep{feng2025llmdroolsmultillmcollaboration}. The eLLM’s capacity for automatic error counterbalancing represents a practical form of self-correction—an essential step toward reliable artificial intelligence.

This paper makes the following contributions:
\begin{itemize}[leftmargin=1.5em]
  \item We formalize an ensemble LLM framework for taxonomy-based categorization, specifying aggregation criteria within a collective decision-making model tailored to sparse hierarchical label spaces.
  \item We instantiate eLLM with multiple committee sizes (2, 3, 5, 7, and 10 models) and evaluate them under uniform zero-shot prompting on a human-annotated corpus of 8{,}660 samples labeled with the IAB~2.2 taxonomy.
  \item We provide empirical evidence of substantial improvements over the best single model, including large gains in F1-score, reduced category inflation, and lower hallucination rates, and we analyze the computational trade-offs of ensembling.
  \item We discuss conditions under which eLLM approaches human-expert performance, and the implications for scalable, reliable labeling pipelines.
\end{itemize}

Together, these results suggest that ensemble-based categorization is a practical and robust alternative to single-model LLM classification, particularly when adherence to a fixed taxonomy and reliability under distributional shift are required. This work builds toward a new theory of collaborative AI, where orchestrated ensembles transform unstructured chaos into ordered, expert-level insight.

\section{Related Work}\label{sec:related-work}

Our research on ensemble large language models (eLLM) for hierarchical text categorization builds upon and integrates advances in three major domains: (1) taxonomy-based classification, (2) zero-shot categorization with large language models, and (3) collective decision-making and ensemble learning. It directly extends the findings of \citep{kamen2025order}, which identified structural limitations in single-model LLM categorization and established the empirical foundation for ensemble-based approaches.

\subsection{Taxonomy-Based Hierarchical Classification}

Text categorization has been a central task in natural language processing (NLP) for several decades. Early systems relied on manual annotation and rule-based heuristics, later replaced by supervised pipelines built upon feature engineering and linear classifiers \citep{Sebastiani2002}. As taxonomies expanded in depth and scope, hierarchical classification emerged as a natural extension, in which labels are organized into a tree or directed acyclic graph \citep{SillaFreitas2011Survey}.

Hierarchical classification requires predictions that respect structural dependencies, maintain parent--child consistency, and remain valid across all levels of granularity. Various methods have been proposed to address these challenges, including structural loss functions \citep{Hsu2009Multiclass}, regularization \citep{Ramaswamy2015HierarchicalLoss}, and post-hoc constraint enforcement \citep{Kosmopoulos2015Evaluation}. In applied frameworks such as the Interactive Advertising Bureau (IAB) taxonomy, label ambiguity and uneven branch depth present additional difficulties. Hierarchical regularization \citep{CesaBianchi2006Hierarchical} and ontology-based inference have been explored as mitigation strategies, though both remain dependent on domain-specific supervision.

\subsection{Zero-Shot Classification and Limitations of Individual LLMs}

Pre-trained and instruction-tuned large language models (LLMs) have popularized zero-shot and few-shot classification via prompt-based label verbalization \citep{Brown2020, Raffel2020, Ouyang2022RLHF}. By aligning textual content and label descriptions within a shared linguistic space, LLMs can perform taxonomy mapping without task-specific fine-tuning. Despite their generality, however, single LLMs display instability and reduced reliability when tasked with mapping unstructured text to sparse taxonomies.

Empirical studies have reported sensitivity to prompt phrasing, seed variation, and deviations from ontology constraints \citep{xu-etal-2024-classification, sun-etal-2023-carp}. Moreover, hallucination and category inflation---the tendencies to generate non-existent or excessive labels---remain recurrent issues in taxonomy-based classification \citep{Ji2022HallucinationSurvey}. \citep{kamen2025order} performed a comparative analysis of ten LLMs under uniform zero-shot conditions on a human-annotated corpus of 8{,}660 samples labeled with the IAB~2.2 taxonomy, demonstrating a consistent performance plateau across architectures, parameter counts, and model generations. These results support the view that text categorization is fundamentally an information-compression problem that requires structured external grounding; ensemble strategies help compensate by aggregating diverse model perspectives. Instruction tuning contributes to robustness but does not eliminate these limitations \citep{Mishra2022InstructionTuning}.

\subsection{Ensembles and Collective Decision-Making}

Ensembling is a well-established method for improving predictive stability and generalization by combining outputs from multiple learners \citep{Breiman1996Bagging, kuncheva2004combining}. Within the LLM domain, ensemble-like mechanisms have appeared in three main forms:

\textit{Internal aggregation.} Self-consistency \citep{Wang2023SelfConsistency} improves reasoning by sampling multiple responses from a single model and aggregating them to reduce variance. Similarly, Mixture-of-Experts (MoE) architectures distribute computation across specialized subnetworks to achieve internal diversity \citep{Fedus2022Switch}.

\textit{Multi-agent collaboration.} Debate-based and critique-based frameworks leverage interaction among models to refine outputs and correct factual or logical errors \citep{Du2023Debate, Madaan2023SelfRefine}.

Our approach differs by formalizing external collective decision-making (CDM) among heterogeneous, black-box LLMs. This draws on ensemble theory and statistical methods for aggregating human annotations \citep{DawidSkene1979EM, Snow2008NonExpert}. Unlike within-model ensembles or debate systems, eLLM integrates independent model judgments under explicit aggregation rules, ensuring hierarchical validity while suppressing common error modes such as hallucination and off-ontology labeling. The CDM formulation treats each LLM as a bounded rational agent and defines decision criteria---such as majority, weighted consensus, or quorum-based voting---optimized for structured label spaces. Related work on controlled diversity in one-shot ensembles parallels our emphasis on balancing stability and variety to manage sparse taxonomic mappings \citep{niimi2025diversity}.

\subsection{Theoretical Foundations of Collective Intelligence}

The theoretical basis for ensemble decision-making traces to early studies of collective intelligence and voting theory. The Condorcet Jury Theorem established that, under the assumption of individual competence above chance and independent errors, the probability of a correct majority decision increases with group size \citep{Condorcet1785Essai}. In machine learning, ensemble variance reduction and bias correction are grounded in similar probabilistic reasoning \citep{Breiman1996Bagging, FreundSchapire1997Boosting, Wolpert1992StackedGeneralization}. In human and hybrid systems, collective decision frameworks weight contributors based on reliability or confidence \citep{Raykar2010Crowds, Welinder2010Crowds}. The eLLM framework adapts these principles to large language models, where each model behaves as a stochastic classifier with an implicit knowledge prior. The ensemble thus functions as a distributed decision system whose
accuracy derives from both model diversity and structured aggregation.

This connection provides theoretical justification for eLLM’s superior performance: if
individual models exceed random accuracy and exhibit partially independent error patterns,
the ensemble’s majority decision probability converges toward human-expert levels.
Moreover, aggregation rules incorporating abstention, confidence calibration, or quorum
thresholds enable eLLM to approximate Bayesian-optimal decision fusion under
uncertainty—enhancing performance on multi-relevant, sparsely annotated categorizations

\subsection{Positioning and Contribution}

Our research unifies hierarchical classification, zero-shot LLM limitations, and collective intelligence theory into a single ensemble framework. While prior ensemble strategies for LLMs often relied on sampling variation or intra-model routing, eLLM performs decision-level aggregation across independently trained models of diverse architectures and release generations. Building upon the empirical findings of \citep{kamen2025order}, which defined a performance ceiling for individual LLMs, this study demonstrates that structured aggregation across model families can consistently exceed that threshold. The proposed eLLM delivers measurable improvements in accuracy, precision, recall, F1-score, reliability, and robustness, offering a scalable paradigm for ensemble reasoning in contexts requiring interpretability, stability, and adherence to external ontologies.

\section{Mathematical Model: Categorization, Collective Decision Making (CDM) Criteria and Framework, Performance Analysis}\label{sec:mathmodel}

The ensemble categorization framework consists of five primary modules: (1) categorization, (2) expert labeling of the test data, (3) ensemble pool construction, (4) CDM criteria and algorithms, and (5) performance management. The following mathematical formulation describes and formalizes these components.

\subsection{Categorization Procedure}

Let $T$ denote a hierarchical taxonomy of categories. The categorization procedure $C$ applied to a text $x$ generates a (set-valued) assignment:
\begin{equation}
C(T,x)=\{t_1,t_2,\ldots,t_n\}.
\tag{3.1}\label{eq:3.1}
\end{equation}
Each generated category in \eqref{eq:3.1} may either belong to the taxonomy $T$ or represent a hallucination (i.e., any string not contained in $T$). When the categorization procedure is a zero-shot large language model, we denote it as $\mathrm{LLM}(T,x)$. The order of elements in \eqref{eq:3.1} is irrelevant.

\subsection{Expert Categorization}

To evaluate any categorization procedure, a benchmark reference is defined. Let $E(T,x)$ denote the expert (human or verified ground-truth) categorization of text $x$:
\begin{equation}
E(T,x)=\{e_1,e_2,\ldots,e_m\},\qquad e_j\in T,\; e_i\neq e_j.
\tag{3.2}\label{eq:3.2}
\end{equation}
Expert categorization uses only valid categories from $T$, all of which are relevant and characterized by maximal accuracy and precision.

\subsection{Categorization Ensembles and CDM Algorithms}

Assume $k$ independent categorization procedures. Each produces a unique set of categories:
\begin{equation}
C_1(T,x)=\{t_{11},\ldots,t_{1n_1}\},\;\; C_2(T,x)=\{t_{21},\ldots,t_{2n_2}\},\;\ldots,\; C_k(T,x)=\{t_{k1},\ldots,t_{k n_k}\}.
\tag{3.3}\label{eq:3.3}
\end{equation}
When each categorization procedure corresponds to a separate LLM:
\begin{equation}
\mathrm{LLM}_1(T,x),\;\mathrm{LLM}_2(T,x),\;\ldots,\;\mathrm{LLM}_k(T,x)\;\longrightarrow\; C_1,\;C_2,\;\ldots,\;C_k.
\tag{3.4}\label{eq:3.4}
\end{equation}
The CDM algorithm aggregates these $k$ independent results into a single consensus decision:
\begin{equation}
\mathrm{CDM}(C_1,C_2,\ldots,C_k;T,x)=\{t_1,t_2,\ldots,t_p\},\quad t_j\in T,\; t_i\neq t_j.
\tag{3.5}\label{eq:3.5}
\end{equation}
Equivalently, for LLM ensembles:
\begin{equation}
\mathrm{CDM}(\mathrm{LLM}_1,\mathrm{LLM}_2,\ldots,\mathrm{LLM}_k;T,x)=\{t_1,t_2,\ldots,t_p\},\quad t_j\in T.
\tag{3.6}\label{eq:3.6}
\end{equation}
In this study, all LLMs are treated as independent categorization entities, following principles of expert collective behavior \citep{Arrow1951, Moulin1988}. Modern CDM approaches for multi-LLM ensembles apply voting, scoring, and ranking mechanisms to robustly select the most relevant categories from diverse model outputs \citep{tran2025multiagent}.

\subsection{Relevance Score CDM Criterion}

The CDM framework assigns each category $c$ a \emph{Category Relevance} score $R(c)$, quantifying its overall importance and consensus within the ensemble:
\begin{equation}
R(c)=G(Q_c,I_c,\mathrm{Prox}_c),\qquad 0\le R(c)\le 1.
\tag{3.7}\label{eq:3.7}
\end{equation}
Common formulations include the multiplicative, arithmetic-mean, and geometric-mean models:
\begin{equation}
R(c)=Q_c\cdot I_c\cdot \mathrm{Prox}_c,
\tag{3.8}\label{eq:3.8}
\end{equation}
\begin{equation}
R(c)=\frac{Q_c+I_c+\mathrm{Prox}_c}{3},
\tag{3.9}\label{eq:3.9}
\end{equation}
\begin{equation}
R(c)=\bigl(Q_c\cdot I_c\cdot \mathrm{Prox}_c\bigr)^{\!1/3}.
\tag{3.10}\label{eq:3.10}
\end{equation}

\subsubsection{Category Popularity.}
Popularity measures how often category $c$ appears across the ensemble of size $N$:
\begin{equation}
Q_c=\frac{N_c}{N},
\tag{3.11}\label{eq:3.11}
\end{equation}
where $N_c$ is the number of ensemble members selecting category $c$.

\subsubsection{Category Importance.}
Importance $I_c$ is defined as a function of hierarchical depth:
\begin{equation}
I_c = G_I\!\left(\frac{L_c}{L_{\max}}\right),
\tag{3.12}\label{eq:3.12}
\end{equation}
and, in the simplest case,
\begin{equation}
I_c=\frac{L_c}{L_{\max}},
\tag{3.13}\label{eq:3.13}
\end{equation}
where $L_c$ is the level of category $c$ and $L_{\max}$ is the maximum depth.

\subsubsection{Category Proximity.}
The proximity score $\mathrm{Prox}_c$ quantifies how closely a category $c$ relates to other clusters $C_1,\ldots,C_{k-1}$ formed within the ensemble outputs. Define $\mathrm{prox}(c,C_i)=\max_{c_{j,i}\in C_i}\mathrm{prox}(c,c_{j,i})$. Then
\begin{equation}
\mathrm{Prox}_c=\frac{1}{k-1}\sum_{i=1}^{k-1}\mathrm{prox}(c,C_i).
\tag{3.14}\label{eq:3.14}
\end{equation}

\subsection{Combined Relevance Criterion}

By substitution, the combined relevance can be written as
\begin{equation}
R(c)=G\!\left(\frac{N_c}{N},\,\frac{L_c}{L_{\max}},\,\mathrm{Prox}_c\right),
\tag{3.15}\label{eq:3.15}
\end{equation}
and, in the simplest multiplicative case,
\begin{equation}
R(c)=\frac{N_c\,L_c\,\mathrm{Prox}_c}{N\,L_{\max}}.
\tag{3.16}\label{eq:3.16}
\end{equation}

\subsection{Consensus Threshold}

The \emph{Consensus Threshold} $\tau$ in the CDM is an empirically derived value based on human-annotated categorization benchmarks. Only categories with $R(c)\ge \tau$ are preserved in the final consensus output. Calibration of $\tau$ balances precision and recall and is reported in Section~\ref{sec:performance-eval} (see threshold sweeps and ablations).

\section{LLM Evaluation Criteria}\label{sec:llm-eval}

To evaluate categorization performance, we define a unified framework that includes both classical metrics and LLM-specific measures. This section formalizes the evaluation methodology used for both single-model and ensemble experiments.

\subsection{Fundamental Definitions}

Let each categorization procedure produce a set of predicted categories compared against an expert-annotated benchmark. We define:
\begin{itemize}[leftmargin=1.5em]
  \item \textbf{True Positives (TP):} Categories assigned by both the LLM and the expert.
  \item \textbf{False Positives (FP):} Categories predicted by the LLM but absent from the expert set.
  \item \textbf{False Negatives (FN):} Categories assigned by the expert but missed by the LLM.
\end{itemize}

\subsection{Classic Evaluation Criteria}\label{sec:classic-eval}

The four standard performance metrics are defined as follows:

\begin{itemize}[leftmargin=1.5em]
    \item \textbf{Accuracy:} The overall proportion of correct predictions.
\end{itemize}

\begin{equation}
\text{Accuracy} = \frac{TP}{TP + FP + FN}
\tag{4.1}\label{eq:4.1}
\end{equation}

\begin{itemize}[leftmargin=1.5em]
    \item \textbf{Precision:} The fraction of predicted categories that are correct.
\end{itemize}

\begin{equation}
\text{Precision} = \frac{TP}{TP + FP}
\tag{4.2}\label{eq:4.2}
\end{equation}

\begin{itemize}[leftmargin=1.5em]
    \item \textbf{Recall:} The fraction of expert categories recovered by the model.
\end{itemize}

\begin{equation}
\text{Recall} = \frac{TP}{TP + FN}
\tag{4.3}\label{eq:4.3}
\end{equation}

\begin{itemize}[leftmargin=1.5em]
    \item \textbf{F1-score (F1):} The harmonic mean of Precision and Recall, balancing both metrics into a single measure.
\end{itemize}

\begin{equation}
F1 = 2 \times \frac{(\text{Precision} \times \text{Recall})}{\text{Precision} + \text{Recall}}
\tag{4.4}\label{eq:4.4}
\end{equation}

\subsection{Additional Criteria for LLM-Based Categorization}

While traditional metrics measure correctness, they do not fully capture behaviors specific to LLMs. We therefore introduce three complementary criteria: Hallucination Ratio (HR), Category Inflation Ratio (IR), and Computation Cost (Cost).

\paragraph{Hallucination Ratio (HR).}
Let
\begin{equation}
h_x=\{\,t_i\in \mathrm{LLM}(T,x)\;:\; t_i\notin T\,\}.
\tag{4.5}\label{eq:4.5}
\end{equation}
Then the Hallucination Ratio for sample $x$ is
\begin{equation}
\mathrm{HR}_x=\frac{\lvert h_x\rvert}{\lvert \mathrm{LLM}(T,x)\rvert}.
\tag{4.6}\label{eq:4.6}
\end{equation}

\paragraph{Category Inflation Ratio (IR).}
The relative expansion of predicted labels compared to expert annotations is
\begin{equation}
\mathrm{IR}_x=\frac{\lvert \mathrm{LLM}(T,x)\rvert}{\lvert E(T,x)\rvert}.
\tag{4.7}\label{eq:4.7}
\end{equation}
To mitigate redundancy in hierarchical taxonomies, we apply the Parent Exclusion Rule (PER), which removes a parent if its child is present. The reduced inflation ratio is
\begin{equation}
\mathrm{IR}^{\ast}_x=\frac{\lvert \mathrm{PER}(\mathrm{LLM}(T,x))\rvert}{\lvert E(T,x)\rvert}.
\tag{4.8}\label{eq:4.8}
\end{equation}

\paragraph{Price of Computation (Cost).}
For each input text $x$, the monetary expense associated with token processing is
\begin{equation}
\mathrm{Cost}_x=(c_{\mathrm{in}}\times N_{\mathrm{in}}(x))+(c_{\mathrm{out}}\times N_{\mathrm{out}}(x)),
\tag{4.9}\label{eq:4.9}
\end{equation}
where $c_{\mathrm{in}}$ and $c_{\mathrm{out}}$ are the prices per input and output token, and $N_{\mathrm{in}}(x)$, $N_{\mathrm{out}}(x)$ are the respective input and output token counts.

\subsection{Corpus-level Aggregation}

For a corpus of texts $D=\{x_1,x_2,\ldots,x_k\}$, metrics are averaged as
\begin{equation}
\mathrm{Metric}(D) = \frac{1}{k} \sum_{i=1}^{k} \mathrm{Metric}(x_i)
\tag{4.10}\label{eq:4.10}
\end{equation}

Total computational cost is calculated as

\begin{equation}
\mathrm{TotalCost}(D) = \sum_{i=1}^{k} \mathrm{Cost}(x_i)
\tag{4.11}\label{eq:4.11}
\end{equation}

Traditional metrics such as accuracy, precision, recall, and F1 provide a well-established foundation for evaluating correctness. However, LLMs introduce distinct challenges, including hallucinations (generation of invalid categories), inflation (over-assignment of categories), and high computational cost due to token-based inference. By incorporating HR, IR, and Cost, this framework captures both the qualitative and quantitative dimensions of performance—offering a more realistic assessment of LLM-based categorization at scale. This combination of classical and novel measures allows evaluation not only of correctness but also of reliability, efficiency, and practical deployability. To demonstrate the practical operation of the ensemble’s collective decision-making (CDM) framework, we now present a concrete example that illustrates how multiple LLMs jointly resolve category assignments for a representative text.

\section{Experimental Setup and Evaluation}\label{sec:experimental-setup}

The experimental setup for evaluating the LLM ensemble performance employed a dataset of 8{,}660 expert-categorized unstructured texts and ten large language models described in earlier sections. The categorization workflow consisted of four stages: data collection, expert labeling, individual LLM categorization using structured prompts, and ensemble post-processing via the Collective Decision-Making (CDM) algorithm. 

For the ensemble experiments, outputs from all participating LLMs were combined, and the CDM algorithm was applied to compute category relevance scores and filter results according to the consensus threshold defined in Section~\ref{sec:mathmodel}. This ensured that only categories supported by sufficient inter-model agreement were included in the final ensemble predictions.

\subsection{Evaluation Taxonomy and Prompt Design}

The Interactive Advertising Bureau (IAB) taxonomy was selected as the evaluation framework due to its widespread adoption in the online advertising and content classification domains. The IAB~2.2 taxonomy comprises 698 general-purpose categories organized across four hierarchical levels, representing one of the most comprehensive and industry-standard taxonomies available for large-scale content classification.

\begin{figure}[h]
    \centering
    \includegraphics[width=0.9\linewidth]{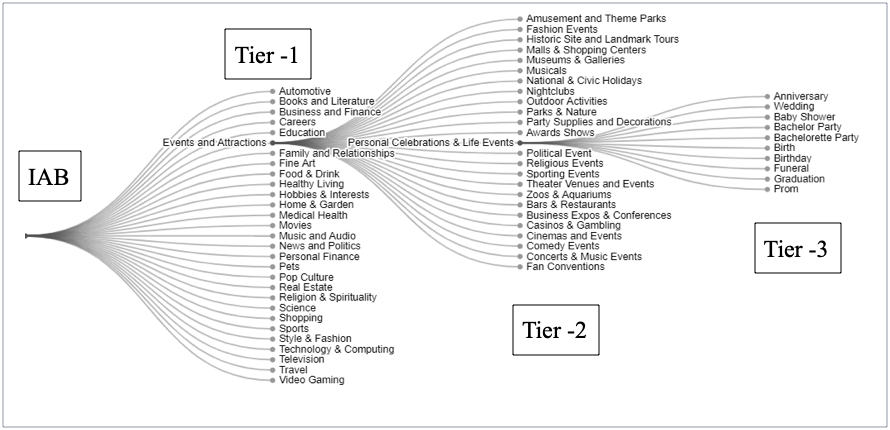}
    \caption{Overview of the IAB Taxonomy}
    \label{fig:iab-taxonomy}
\end{figure}

Figure \ref{fig:iab-taxonomy} illustrates the overall hierarchical structure of the IAB taxonomy.

Each large language model was instructed to first generate a Tier-1 category and then progressively refine the classification through Tier-2, Tier-3, and optionally Tier-4 prompts. This stepwise prompting procedure mirrors human reasoning by narrowing the categorization scope from general to specific domains.

All models were evaluated under uniform zero-shot conditions using the same prompt structure and taxonomy constraints. System prompts defined the classification task and enforced compliance with the IAB category list, while user prompts contained the target text. To minimize hallucinations and ensure taxonomic validity, allowable categories and structural rules were embedded directly within the system prompt \citep{kamen2025order}.

\section{Example of CDM Computation}\label{sec:cdm-example}

To illustrate the application of the ensemble CDM framework for unstructured text categorization, we analyze a randomly selected passage from the dataset:

\begin{quote}
I turned my attention to Braw. “Tomorrow night, we will have a small demonstration of your power. You will help me liberate five beings from their demons. If all goes well, we will wait four nights until the moons align with the star of Juns and when the magic of this world is at its peak, we will channel the power of Laif through the entire world, whereupon we will rid the world of demons.” My jaw dropped. “You’ll kill them all!” Braw shook his head. “No dear. We will free them.” “You don’t get it!” I could feel panic taking a hold of me. “The Namaels, the Majs, they’re not possessed! That’s just who they are! It’s genetic!” I looked up at Brice and was shocked by the blank look on his face. “Brice! You have to explain to him. It will kill you!” Brice turned slowly to look at me, his eyes wide. Hayden has made up her mind. If she doesn’t pick sides between the Namaels and the Majs then innocent beings won’t die because of her. Of course, things can never be that simple and when she is taken prisoner by the humans a whole new problem arises. It doesn’t take long before everything she thought she knew no longer makes sense. Could it be possible that the prophecy was wrong?
\end{quote}

Ten LLMs (Claude~3.5, Gemini~1.5, Gemini~2.0, Llama~3~8B, Llama~3.3~70B, DeepSeek, xAI~Grok, Mistral, GPT~OSS-20B, and GPT~OSS-120B) each produced an independent set of generic IAB~v2.2 categories for this text. The individual outputs highlight both the diversity and the occasional disagreement inherent in model interpretations.

\begin{table}[h]
    \centering
    \footnotesize
    \setlength{\tabcolsep}{2.5pt} 
    \renewcommand{\arraystretch}{1.2} 
    \begin{tabular}{|p{1.6cm}|p{1.6cm}|p{1.6cm}|p{1.6cm}|p{1.6cm}|p{1.6cm}|p{1.6cm}|p{1.6cm}|p{1.6cm}|p{1.6cm}|}
        \hline
        \textbf{Claude} & \textbf{Gemini 1.5} & \textbf{Gemini 2.0} & \textbf{Llama 3 8B} & \textbf{Llama 3.3 70B} & \textbf{DeepSeek} & \textbf{xAI Grok} & \textbf{Mistral} & \textbf{GPT OSS-20B} & \textbf{GPT OSS-120B} \\
        \hline
        Books and Literature, Fiction, Young Adult Literature & 
        Books and Literature, Fiction & 
        Books and Literature, Young Adult Literature, Fiction & 
        Books and Literature, Fantasy, Pop Culture, Fiction & 
        Books and Literature, Pop Culture, Fiction, Young Adult Literature & 
        Books and Literature, Fiction, Young Adult Literature & 
        Religion \& Spirituality, Sensitive Topics & 
        Books and Literature, Science, Fiction, Young Adult Literature, Genetics &
        Books and Literature, Fiction &
        Books and Literature, Fiction, Young Adult Literature \\
        \hline
    \end{tabular}
    \caption{IAB Categorizations of Sample Text by 10 LLMs}
    \label{tab:llm_categorizations}
\end{table}

Applying the collective decision-making mechanism defined in Equation~\eqref{eq:3.16} and using a consensus threshold $\tau = 0.65$ (empirically calibrated during evaluation), the ensemble aggregates individual predictions into a unified result.

The resulting relevance scores for the top-ranked categories are:
\begin{itemize}[leftmargin=1.5em]
  \item Fiction. $R_C = 1.00$
  \item Young Adult Literature. $R_C = 0.79$
\end{itemize}

The ensemble’s final output, therefore, is:
\[
E\text{-LLM Prediction} = \{\text{Fiction, Young Adult Literature}\}.
\]

This outcome achieves a 100~percent match with the human expert benchmark:
\[
\text{Expert Categorization} = \{\text{Fiction, Young Adult Literature}\}.
\]

The example demonstrates how the eLLM framework consolidates diverse model judgments into a consistent, taxonomy-valid, and semantically robust decision. It also shows that ensemble consensus not only suppresses hallucinated or irrelevant categories but reproduces human-expert precision without the need for retraining or fine-tuning.

\section{Performance Evaluation and Comparative Results}\label{sec:performance-eval}

The experimental results demonstrate the significant performance gains achieved by the ensemble large language model (eLLM) categorization framework. The following subsections analyze performance across confidence thresholds, ensemble sizes, and model compositions, and compare ensemble results against individual LLMs.

\subsection{Ensemble Threshold Performance Analysis} 

The Collective Decision-Making (CDM) criterion was evaluated using nine confidence thresholds 
\[
T = \{0.55, 0.60, 0.65, 0.70, 0.75, 0.80, 0.85, 0.90, 0.95\}
\]
on the test corpus of 8{,}660 documents. The results are presented in Table~\ref{tab:threshold-performance}.

\begin{table}[htbp]
\centering
\caption{Performance Metrics Across Consensus Thresholds}
\label{tab:threshold-performance}
\begin{tabular}{lcccccccccc}
\toprule
\textbf{Metric} & 0.50 & 0.55 & 0.60 & 0.65 & 0.70 & 0.75 & 0.80 & 0.85 & 0.90 & 0.95 \\
\midrule
F1 & 0.78 & 0.83 & 0.88 & 0.92 & 0.85 & 0.81 & 0.74 & 0.69 & 0.63 & 0.58 \\
Accuracy & 0.69 & 0.78 & 0.86 & 0.94 & 0.97 & 1.00 & 1.00 & 1.00 & 1.00 & 1.00 \\
Precision & 0.69 & 0.78 & 0.86 & 0.94 & 0.97 & 1.00 & 1.00 & 1.00 & 1.00 & 1.00 \\
Recall & 1.00 & 0.97 & 0.95 & 0.93 & 0.81 & 0.72 & 0.64 & 0.58 & 0.52 & 0.45 \\
\bottomrule
\end{tabular}
\end{table}

Threshold $T = 0.65$ achieved the optimal trade-off between precision, recall, and F1 score. This threshold was therefore selected as the default confidence parameter for ensemble-based categorization.

\subsection{F1 Score Improvement by Ensemble Size}

Table~\ref{tab:f1-ensemble-size} summarizes how F1 scores increase with ensemble size, showing consistent performance gains as additional LLMs are incorporated.

\begin{table}[htbp]
\centering
\caption{F1-score (F1) Improvement by Ensemble Size}
\label{tab:f1-ensemble-size}
\begin{tabular}{lcc}
\toprule
\textbf{Ensemble Configuration} & \textbf{F1} & \textbf{F1 Improvement (\%)} \\
\midrule
Single LLM (best) & 0.55 & -- \\
2 LLMs (Gemini 2.0, DeepSeek) & 0.73 & +33\% \\
3 LLMs (Gemini 2.0, Llame 3.3 70B, DeepSeek) & 0.76 & +38\% \\
10 LLMs & 0.92 & +67\% \\
\bottomrule
\end{tabular}
\end{table}

These results confirm that ensemble-based categorization substantially outperforms individual LLMs, particularly as ensemble diversity increases.

\subsection{Detailed Comparison by Ensemble Composition}

To further explore the relationship between ensemble composition and performance, Table~\ref{tab:ensemble_composition} details F1 scores for different model combinations and ensemble sizes.

\begin{table}[h!]
\scriptsize
\centering
\caption{F1 Score by Ensemble Composition}
\label{tab:ensemble_composition}
\begin{tabular}{cll}
\toprule
\textbf{Ensemble Size} & \textbf{F1 Score} & \textbf{Ensemble LLMs} \\
\midrule
1 & 0.53 & Gemini 2.0 \\
1 & 0.51 & Llame 3.3 70B \\
1 & 0.52 & DeepSeek \\
2 & 0.73 & Gemini 2.0, DeepSeek \\
2 & 0.70 & Gemini 2.0, Llame 3.3 70B \\
3 & 0.76 & Gemini 2.0, Llame 3.3 70B, DeepSeek \\
3 & 0.76 & Gemini 2.0, Claude, Llame 3.3 70B \\
4 & 0.79 & Gemini 2.0, Claude, Llame 3.3 70B, DeepSeek \\
5 & 0.82 & Gemini 2.0, Claude, Llame 3.3 70B, DeepSeek, Grok \\
5 & 0.82 & Gemini 2.0, Claude, Llame 3.3 70B, DeepSeek, Mistral \\
7 & 0.85 & Gemini 2.0, Claude, Llame 3.3 70B, DeepSeek, Mistral, Grok, Llama 3 8B \\
7 & 0.85 & Gemini 2.0, Claude, Llame 3.3 70B, DeepSeek, Mistral, Grok, Gemini 1.5 \\
8 & 0.87 & Gemini 2.0, Claude, Llame 3.3 70B, DeepSeek, Mistral, Grok, Gemini 1.5, Llama 3 8B\\
10 & 0.92 & Gemini 2.0, Claude, Llame 3.3 70B, DeepSeek, Mistral, Grok, Gemini 1.5, Llama 3 8B , GPT OSS-20B, GPT OSS-120B \\

\bottomrule
\end{tabular}
\end{table}

\subsection{Comparison to Individual LLMs}

Tables~\ref{tab:f1-ensemble-size} and~\ref{tab:ensemble_composition} show that the ensemble framework consistently surpasses individual LLM performance across all metrics. While the best single model achieved an F1 score of 0.55, even a two-model ensemble reached 0.73. Larger ensembles of five to ten models achieved F1 scores exceeding 0.80 and approaching 0.90, confirming the scalability of collective intelligence in LLM categorization~\citep{kamen2025order}.

To contextualize these results, Table~\ref{tab:individual-llm-performance} presents individual model performance on the full 8{,}660-document dataset. While Claude~3.5 (Claude) attained the highest single-model F1 score, differences among the top-performing models were small. All systems exhibited recall variability, category inflation, and occasional hallucinations. Notably, GPT~OSS-120B maintained the lowest hallucination rate but still plateaued in precision and recall.

Overall, ensemble categorization effectively reduces noise, mitigates hallucinations, and harmonizes inconsistent category use. These improvements come at the cost of increased computation, as multiple models are evaluated per document. However, this cost can be optimized by tuning ensemble size to achieve a balance between accuracy and efficiency.

\subsection{Computation Cost}

Token pricing remains a key factor in ensemble deployment. Table~\ref{tab:pricing-models} summarizes approximate input and output token costs as of September~2025. The wide disparity in pricing—from approximately \$0.04 to \$30.00 per million tokens—underscores the importance of cost-aware ensemble design.

\begin{table}[htbp]
\centering
\caption{Individual LLM Performance on 8{,}660-Sample IAB Dataset}
\label{tab:individual-llm-performance}
\begin{tabular}{lcccc}
\toprule
\textbf{Model} & \textbf{F1} & \textbf{Accuracy} & \textbf{Precision} & \textbf{Recall} \\
\midrule
Claude~3.5 & 0.55 & 0.52 & 0.46 & 0.79 \\
Gemini~1.5 & 0.49 & 0.54 & 0.45 & 0.64 \\
Gemini~2.0~ & 0.52 & 0.54 & 0.46 & 0.72 \\
LLaMA~3~8B & 0.39 & 0.41 & 0.33 & 0.60 \\
LLaMA~3.3~70B & 0.51 & 0.43 & 0.40 & 0.87 \\
DeepSeek & 0.52 & 0.51 & 0.45 & 0.75 \\
Grok & 0.50 & 0.55 & 0.46 & 0.66 \\
Mistral & 0.47 & 0.41 & 0.36 & 0.83 \\
GPT OSS-20B & 0.52 & 0.55 & 0.47 & 0.71 \\
GPT OSS-120B & 0.53 & 0.55 & 0.47 & 0.72 \\
\bottomrule
\end{tabular}
\end{table}

\begin{table}[htbp]
\centering
\caption{LLM Pricing Models as of August~2025 (per 1M tokens)}
\label{tab:pricing-models}
\begin{tabular}{lcc}
\toprule
\textbf{Model} & \textbf{Input Cost} & \textbf{Output Cost} \\
\midrule
Claude~3.5 & \$0.80 & \$4.00 \\
Gemini~1.5~ & \$0.04 & \$0.15 \\
Gemini~2.0~ & \$0.10 & \$0.40 \\
Mistral~(Nemo) & \$8.00 & \$8.00 \\
LlaMA~3~8B~ & \$0.05 & \$0.08 \\
Llama~3.3~70B~ & \$0.59 & \$0.79 \\
Grok & \$2.00 & \$10.00 \\
DeepSeek & \$0.27 & \$1.10 \\
GPT OSS-20B~ & \$0.10 & \$0.50 \\
GPT OSS-120B~ & \$0.15 & \$0.75 \\
\bottomrule
\end{tabular}
\end{table}

While open-source models offer cost advantages, hybrid ensembles that combine both proprietary and open-weight LLMs achieve the best balance between cost and accuracy. The ensemble evaluation confirms that coordinated multi-model decision-making provides measurable advantages over any individual LLM. Having established the quantitative evidence of these improvements, we now examine the broader implications, limitations, and theoretical significance of ensemble-based categorization.

\section{Discussion}\label{sec:discussion}

Given the rapid advances and increasing power of contemporary large language models, one might expect substantially higher accuracy, precision, recall, and F1 scores. However, zero-shot experiments continue to produce only moderate results, revealing a persistent gap between general generative competence and the targeted classification of unstructured text within a complex hierarchical taxonomy.

Categorization is a uniquely constrained task that requires compressing rich, unstructured content into a sparse, predefined label space. Although the Interactive Advertising Bureau (IAB) taxonomy is a well-designed and widely adopted framework, it includes only 690 general-purpose categories—insufficient to capture the full complexity and diversity of real-world content. This limitation becomes particularly evident when compared to the DMOZ taxonomy (the Open Directory Project), which comprises more than 750,000 nodes across multiple hierarchical levels. Achieving human-expert-level IAB categorization therefore demands not only linguistic fluency but also a taxonomy-aware world model capable of consolidating dispersed semantic evidence into concise, non-overlapping labels. Scaling and architectural improvements alone do not guarantee better performance. Newer or larger models—such as Gemini~2.0 or GPT~OSS-120B—did not consistently outperform their predecessors across classical metrics. For categorization tasks, performance gains appear to diminish beyond a certain model size.

The ensemble paradigm fundamentally changes this dynamic. The eLLM framework significantly enhances categorization quality, achieving over 60~percent improvement in F1 score by effectively filtering low-relevance and misassigned categories while eliminating hallucinations. This demonstrates that ensembling provides an efficient mechanism for overcoming the structural limitations of individual LLMs. Even small ensembles—comprising two to three models—yield substantial quality gains, with F1 improvements exceeding 30~percent. Based on experimental evidence, ensemble-based categorization achieves the highest precision and robustness among available zero-shot methods, making it a preferred strategy for complex classification tasks.

These findings are particularly notable given the ongoing investments in large-scale, standalone LLMs. The results support the assertion that in artificial intelligence–based categorization, majority consensus among models—rather than individual scale—drives superior accuracy. Another critical insight is that collective decision-making algorithms applied to sufficiently large ensembles can achieve categorization performance comparable to, or even surpassing, that of human experts. The prospect of ensemble LLMs consistently exceeding human accuracy has profound implications: it marks a conceptual shift from viewing AI as a ``tool'' to recognizing it as a ``super-expert.''

If ensemble systems can reliably outperform humans in structured classification tasks such as IAB taxonomy mapping, they may enable full automation in traditionally manual domains such as programmatic advertising, content moderation, research indexing, and regulatory compliance. Such automation promises substantial improvements in scalability, cost-efficiency, and consistency—while minimizing risks introduced by human subjectivity or error.

The primary limitation of ensemble categorization lies in computational expense, as it requires multiple parallel API calls. However, this drawback is expected to diminish as token-processing costs continue to decline. This economic trend will further encourage adoption and democratize access to ensemble-based AI solutions. Future research directions may include integrating structural and governance models from multi-agent systems, allowing ensembles to function as dynamic communities of specialized LLMs with distributed decision mechanisms~\citep{ferreira2024society}. Such architectures could enable adaptive weighting, self-regulation, and continuous learning across heterogeneous model ecosystems, moving closer to the vision of collaborative artificial intelligence.

\section{Conclusion}\label{sec:conclusion}

This study demonstrates the superior accuracy, stability, and interpretability of ensemble-based large language model (LLM) categorization compared to individual models. The proposed Collective Decision-Making (CDM) framework effectively mitigates hallucinations, reduces category inflation, and enhances overall categorization precision. Empirical evaluation confirms that majority-based consensus among diverse models produces more reliable results than any single-model approach, reinforcing the broader principle that coordinated orchestration—rather than scale alone—drives performance improvements in large-scale AI systems.

\subsection*{Practical Implications}

Ensemble-based LLM categorization represents a major advancement in the classification of unstructured text. In high-precision domains such as programmatic advertising, academic indexing, legal and policy classification, and regulatory compliance, ensemble frameworks offer a viable alternative to manual annotation. The ability of multi-model ensembles to match or exceed human-level categorization accuracy suggests a path toward scalable, fully automated labeling pipelines.

Organizations that rely on expert review or crowdsourced labeling stand to benefit substantially from ensemble adoption. The combination of higher throughput, lower operational cost, and consistent interpretability makes eLLM systems especially attractive for industrial-scale content management and data governance.

\subsection*{Future Work}

Several research directions can further improve ensemble-based categorization:
\begin{itemize}[leftmargin=1.5em]
    \item \textbf{Dynamic Ensemble Composition:} Development of adaptive ensembles that select LLMs based on domain specialization, contextual relevance, or task complexity.
    \item \textbf{Cost–Performance Optimization:} Identification of the minimal ensemble size and composition that achieves near-maximal accuracy while minimizing inference cost.
    \item \textbf{Enhanced Proximity Metrics:} Refinement of semantic similarity measures to better integrate low-agreement but semantically valid categories.
    \item \textbf{Model Weighting and Fine-Tuning:} Implementation of weighted voting or confidence calibration, assigning greater influence to models with demonstrated reliability on specific domains.
\end{itemize}

As inference costs continue to decline and multi-modal capabilities expand, ensemble methods are expected to become increasingly accessible and deployable at scale. Their integration into autonomous content classification, moderation, and discovery systems will enable more transparent, reliable, and human-aligned AI ecosystems.

The findings of this research suggest that ensemble-based reasoning represents a natural evolution in large language model design. By combining the strengths of multiple architectures within a unified decision framework, eLLM systems achieve both the precision of expert judgment and the scalability of automated processing. This convergence of collaboration and computation marks a decisive step toward the realization of truly collective artificial intelligence.

\bibliographystyle{plainnat}
\bibliography{references}

\begin{thebibliography}{33}
\providecommand{\natexlab}[1]{#1}
\providecommand{\url}[1]{\texttt{#1}}
\expandafter\ifx\csname urlstyle\endcsname\relax
  \providecommand{\doi}[1]{doi: #1}\else
  \providecommand{\doi}{doi: \begingroup \urlstyle{rm}\Url}\fi

\bibitem[Arrow(1951)]{Arrow1951}
K.~J. Arrow.
\newblock \emph{Social Choice and Individual Values}.
\newblock Wiley, 1951.

\bibitem[Breiman(1996)]{Breiman1996Bagging}
Leo Breiman.
\newblock Bagging predictors.
\newblock \emph{Machine Learning}, 24\penalty0 (2):\penalty0 123--140, 1996.
\newblock \doi{10.1023/A:1018054314350}.

\bibitem[Brown et~al.(2020)Brown, Mann, Ryder, et~al.]{Brown2020}
T.~Brown, B.~Mann, N.~Ryder, et~al.
\newblock Language models are few-shot learners.
\newblock \emph{NeurIPS}, 2020.

\bibitem[Cesa-Bianchi et~al.(2006)]{CesaBianchi2006Hierarchical}
N.~Cesa-Bianchi et~al.
\newblock Hierarchical classification: Combining bayes and svm.
\newblock In \emph{Proceedings of the 23rd International Conference on Machine Learning (ICML)}, pages 177--184, 2006.
\newblock \doi{10.1145/1143844.1143868}.

\bibitem[Condorcet(1785)]{Condorcet1785Essai}
M.~de Condorcet.
\newblock \emph{Essai sur l’Application de l’Analyse à la Probabilité des Décisions Rendues à la Pluralité des Voix}.
\newblock Imprimerie Royale, Paris, 1785.

\bibitem[Dawid and Skene(1979)]{DawidSkene1979EM}
A.~P. Dawid and A.~M. Skene.
\newblock Maximum likelihood estimation of observer error-rates using the em algorithm.
\newblock \emph{Applied Statistics}, 28\penalty0 (1):\penalty0 20--28, 1979.
\newblock \doi{10.2307/2346806}.

\bibitem[Du et~al.(2023)]{Du2023Debate}
Yilun Du et~al.
\newblock Improving factuality and reasoning via multi-agent debate.
\newblock \emph{arXiv preprint}, 2023.
\newblock URL \url{https://arxiv.org/abs/2305.11280}.

\bibitem[Fedus et~al.(2022)Fedus, Zoph, and Shazeer]{Fedus2022Switch}
William Fedus, Barret Zoph, and Noam Shazeer.
\newblock Switch transformers: Scaling to trillion parameter models.
\newblock \emph{Journal of Machine Learning Research}, 23\penalty0 (120):\penalty0 1--45, 2022.
\newblock URL \url{https://jmlr.org/papers/v23/21-1045.html}.

\bibitem[Feng et~al.(2025)Feng, Ding, Liu, Wang, Shi, Wang, Shen, Han, Lang, Lee, Pfister, Choi, and Tsvetkov]{feng2025llmdroolsmultillmcollaboration}
Shangbin Feng, Wenxuan Ding, Alisa Liu, Zifeng Wang, Weijia Shi, Yike Wang, Zejiang Shen, Xiaochuang Han, Hunter Lang, Chen-Yu Lee, Tomas Pfister, Yejin Choi, and Yulia Tsvetkov.
\newblock When one llm drools, multi-llm collaboration rules, 2025.
\newblock URL \url{https://arxiv.org/abs/2502.04506}.

\bibitem[Ferreira et~al.(2024)Ferreira, Silva, and Martins]{ferreira2024society}
Silvan Ferreira, Ivanovitch Silva, and Allan Martins.
\newblock Organizing a society of language models: Structures and mechanisms for enhanced collective intelligence.
\newblock \emph{arXiv preprint arXiv:2405.03825}, 2024.
\newblock URL \url{https://arxiv.org/abs/2405.03825}.

\bibitem[Freund and Schapire(1997)]{FreundSchapire1997Boosting}
Yoav Freund and Robert~E. Schapire.
\newblock A decision-theoretic generalization of on-line learning and an application to boosting.
\newblock In \emph{Journal of Computer and System Sciences}, volume~55, pages 119--139, 1997.
\newblock \doi{10.1006/jcss.1997.1504}.

\bibitem[Hsu et~al.(2009)]{Hsu2009Multiclass}
Daniel~J. Hsu et~al.
\newblock Multi-class classification with hierarchical structure.
\newblock In \emph{Proceedings of the 26th International Conference on Machine Learning (ICML)}, pages 361--368, 2009.
\newblock \doi{10.1145/1553374.1553420}.

\bibitem[Ji et~al.(2022)Ji, Lee, Frieske, Yu, Su, Xu, Ishii, Bang, Chen, Dai, Chan, Madotto, and Fung]{Ji2022HallucinationSurvey}
Ziwei Ji, Nayeon Lee, Rita Frieske, Tiezheng Yu, Dan Su, Yan Xu, Etsuko Ishii, Yejin Bang, Delong Chen, Wenliang Dai, Ho~Shu Chan, Andrea Madotto, and Pascale Fung.
\newblock Survey of hallucination in natural language generation.
\newblock \emph{arXiv preprint arXiv:2202.03629}, 2022.
\newblock URL \url{https://arxiv.org/abs/2202.03629}.

\bibitem[Jr. et~al.(2011)Jr., N., and Freitas]{SillaFreitas2011Survey}
Silla Jr., Carlos N., and Alex~A. Freitas.
\newblock A survey of hierarchical classification across different application domains.
\newblock \emph{Data Mining and Knowledge Discovery}, 22\penalty0 (1--2):\penalty0 31--72, 2011.
\newblock \doi{10.1007/s10618-010-0175-9}.

\bibitem[Kamen(2025)]{kamen2025order}
Ariel Kamen.
\newblock Order from chaos: Comparative study of ten leading llms on unstructured data categorization.
\newblock \emph{arXiv preprint arXiv:2510.13885}, 2025.
\newblock URL \url{https://arxiv.org/abs/2510.13885}.

\bibitem[Kosmopoulos et~al.(2015)]{Kosmopoulos2015Evaluation}
Aris Kosmopoulos et~al.
\newblock Evaluation measures for hierarchical classification: A unified view and novel approaches.
\newblock In \emph{Proceedings of the 32nd International Conference on Machine Learning (ICML)}, pages 106--114, 2015.
\newblock URL \url{https://proceedings.mlr.press/v37/kosmo15.pdf}.

\bibitem[Kuncheva(2004)]{kuncheva2004combining}
Ludmila~I. Kuncheva.
\newblock \emph{Combining Pattern Classifiers: Methods and Algorithms}.
\newblock Wiley, 2004.
\newblock \doi{10.1002/0471660264}.
\newblock URL \url{https://onlinelibrary.wiley.com/doi/book/10.1002/0471660264}.

\bibitem[Madaan et~al.(2023)]{Madaan2023SelfRefine}
Ankush Madaan et~al.
\newblock Self-refine: Iterative refinement with feedback.
\newblock In \emph{Advances in Neural Information Processing Systems (NeurIPS)}, 2023.
\newblock URL \url{https://arxiv.org/abs/2303.17651}.

\bibitem[Mishra et~al.(2022)]{Mishra2022InstructionTuning}
Swaroop Mishra et~al.
\newblock Cross-task generalization via instruction tuning.
\newblock In \emph{Advances in Neural Information Processing Systems (NeurIPS)}, 2022.
\newblock URL \url{https://arxiv.org/abs/2104.08773}.

\bibitem[Moulin(1988)]{Moulin1988}
H.~Moulin.
\newblock \emph{Axioms of Cooperative Decision Making}.
\newblock Cambridge University Press, 1988.

\bibitem[Niimi et~al.(2025)Niimi, Kuroda, and Hayashi]{niimi2025diversity}
Yuta Niimi, Rina Kuroda, and Keita Hayashi.
\newblock Controlled diversity in one-shot ensembles for large language models.
\newblock \emph{arXiv preprint arXiv:2505.11234}, 2025.
\newblock TODO: verify metadata.

\bibitem[Ouyang et~al.(2022)]{Ouyang2022RLHF}
Long Ouyang et~al.
\newblock Training language models to follow instructions with human feedback.
\newblock In \emph{Advances in Neural Information Processing Systems (NeurIPS)}, 2022.
\newblock URL \url{https://arxiv.org/abs/2203.02155}.

\bibitem[Raffel et~al.(2020)Raffel, Shazeer, Roberts, , et~al.]{Raffel2020}
C.~Raffel, N.~Shazeer, A.~Roberts, , et~al.
\newblock Exploring the limits of transfer learning.
\newblock \emph{JMLR}, 21\penalty0 (140):\penalty0 1--67, 2020.

\bibitem[Ramaswamy et~al.(2015)]{Ramaswamy2015HierarchicalLoss}
Hema Ramaswamy et~al.
\newblock Hierarchical loss functions for classification.
\newblock In \emph{Proceedings of the 32nd International Conference on Machine Learning (ICML)}, pages 1852--1860, 2015.
\newblock URL \url{https://proceedings.mlr.press/v37/ramaswamy15.pdf}.

\bibitem[Raykar et~al.(2010)]{Raykar2010Crowds}
Vikas~C. Raykar et~al.
\newblock Learning from crowds.
\newblock \emph{Journal of Machine Learning Research}, 11:\penalty0 1297--1322, 2010.
\newblock URL \url{https://www.jmlr.org/papers/volume11/raykar10a/raykar10a.pdf}.

\bibitem[Sebastiani(2002)]{Sebastiani2002}
F.~Sebastiani.
\newblock Machine learning in automated text categorization.
\newblock \emph{ACM Computing Surveys}, 34\penalty0 (1):\penalty0 1--47, 2002.

\bibitem[Snow et~al.(2008)]{Snow2008NonExpert}
Rion Snow et~al.
\newblock Cheap and fast—but is it good? evaluating non-expert annotations for natural language tasks.
\newblock In \emph{Proceedings of the 2008 Conference on Empirical Methods in Natural Language Processing (EMNLP)}, pages 254--263, 2008.
\newblock URL \url{https://aclanthology.org/D08-1027}.

\bibitem[Sun et~al.(2023)Sun, Li, Li, Wu, Guo, Zhang, and Wang]{sun-etal-2023-carp}
Xiaofei Sun, Xiaoya Li, Jiwei Li, Fei Wu, Shangwei Guo, Tianwei Zhang, and Guoyin Wang.
\newblock Text classification via large language models.
\newblock \emph{arXiv preprint}, 2023.
\newblock URL \url{https://arxiv.org/pdf/2305.08377}.

\bibitem[Tran et~al.(2025)Tran, Dao, Nguyen, Pham, O’Sullivan, and Nguyen]{tran2025multiagent}
Khanh-Tung Tran, Dung Dao, Minh-Duong Nguyen, Quoc-Viet Pham, Barry O’Sullivan, and Hoang~D. Nguyen.
\newblock Multi-agent collaboration mechanisms: A survey of llms.
\newblock \emph{arXiv preprint arXiv:2501.06322}, 2025.
\newblock URL \url{https://arxiv.org/abs/2501.06322}.

\bibitem[Wang et~al.(2023)]{Wang2023SelfConsistency}
Xuezhi Wang et~al.
\newblock Self-consistency improves chain of thought in language models.
\newblock \emph{arXiv preprint}, 2023.
\newblock URL \url{https://arxiv.org/abs/2203.11171}.

\bibitem[Welinder et~al.(2010)Welinder, Branson, Perona, and Belongie]{Welinder2010Crowds}
Peter Welinder, Steve Branson, Pietro Perona, and Serge Belongie.
\newblock The multidimensional wisdom of crowds.
\newblock \emph{Advances in Neural Information Processing Systems (NeurIPS)}, 23:\penalty0 2424--2432, 2010.
\newblock URL \url{https://proceedings.neurips.cc/paper/2010/file/cffb3cd0564230e841ee808b7c35b369-Paper.pdf}.

\bibitem[Wolpert(1992)]{Wolpert1992StackedGeneralization}
David~H. Wolpert.
\newblock Stacked generalization.
\newblock \emph{Neural Networks}, 5\penalty0 (2):\penalty0 241--259, 1992.
\newblock \doi{10.1016/S0893-6080(05)80023-1}.

\bibitem[Xu et~al.(2024)Xu, Lou, Du, Mahzoon, Talebianaraki, Zhou, Garrison, Vucetic, and Yin]{xu-etal-2024-classification}
Hanzi Xu, Renze Lou, Jiangshu Du, Vahid Mahzoon, Elmira Talebianaraki, Zhuoan Zhou, Elizabeth Garrison, Slobodan Vucetic, and Wenpeng Yin.
\newblock Llms’ classification performance is overclaimed.
\newblock \emph{arXiv preprint}, 2024.
\newblock URL \url{https://arxiv.org/pdf/2406.16203}.

\end{thebibliography}

\end{document}